\documentclass[runningheads]{llncs}

\usepackage[T1]{fontenc}
\usepackage{lmodern}
\usepackage{graphicx}
\usepackage{booktabs}
\usepackage{amsmath,amssymb}
\usepackage{microtype}
\usepackage{hyperref}
\usepackage[capitalize]{cleveref}
\newcommand{\Rone}{\mathrm{R1}}
\newcommand{\Dmap}{D_{\mathrm{map}}}
\newcommand{\Dpred}{D_{\mathrm{pred}}}

\title{LatticeRank: Spacing-Ranked Consistency for Native-Grid 3D Medical Image Segmentation}
\titlerunning{LatticeRank for Native-Grid 3D Segmentation}
\author{Jingjia Yan\inst{1} \and Yue Zhuo\inst{1} \and
Zhonghua Xie\inst{1}\thanks{Corresponding author: Zhonghua Xie (email: eexie@qq.com).}}
\authorrunning{J. Yan et al.}
\institute{Huizhou University, Huizhou 516007, Guangdong, China\\
\email{1946084073@qq.com, 2937029343@qq.com, eexie@qq.com}}

\begin{document}
\maketitle

\begin{abstract}
Variable voxel spacing changes more than image resolution: it alters the discrete sampling lattice, tensor geometry, and anatomical evidence available to a 3D segmentation network. While conventional pipelines restore images to a canonical grid, direct-grid deployment requires the model itself to tolerate these changes. We observe that candidate grids are not exchangeable views---a coarse grid removes more structural information than a mildly resampled one---yet standard consistency objectives assign them equal influence. We therefore propose LatticeRank, a training-only framework that aligns candidate posteriors in physical space and allocates consistency supervision by voxel-spacing order: among endpoint-matched candidates, the auxiliary loss concentrates on the coarsest requested grid, while inference retains a single network forward with no candidate branch or ensemble. On MSD Task04, LatticeRank improves coarse-grid Extra Dice over mean consistency by 0.03436 across five seeds while meeting the prespecified reference-grid noninferiority criterion. A controlled ladder shows that grid exposure provides the largest recovery, posterior consistency adds a second gain, and spacing-ranked allocation yields the strongest direct-grid result; requested spacing thus provides a simple, model-independent ordering signal for direct-grid consistency learning.
\end{abstract}

\section{Introduction}

Voxel spacing determines the lattice on which anatomy becomes data: increasing through-plane spacing reduces slice count, strengthens partial-volume mixing, and changes the feature sequence presented to a 3D network, so thin structures may disappear before segmentation begins. Existing systems counter this shift through spacing normalization, synthetic resolution diversity, or resolution-aware models~\cite{isensee2021nnunet,billot2023synthseg,henschel2022fastsurfervinn,brudfors2022splatting,joutard2024hyperspace}.

When preprocessing or architecture cannot be changed, some deployments must segment the available tensor, a requirement now shared by speech~\cite{li2026atrie}, multimodal~\cite{zhang2026tokenpenalty,11586066}, and consumer systems~\cite{xu2026mage,11543262} that likewise run on degraded inputs. In this \emph{direct-grid} path, the model inherits both the altered lattice and the information loss of coarse sampling. Interpolation blur is not the whole problem: tensor depth changes, pooling stages operate on a different discrete geometry, and the evidence surviving at 5~mm is not the evidence surviving at 2~mm~\cite{aganj2013interpolation,vanleemput2003partialvolume}. Training on multiple grids exposes the network to this shift but leaves a consequential choice unresolved: how should supervision be divided among candidate spacings sampled in the same update?

The candidates are not exchangeable augmentations: mean consistency gives mildly and severely undersampled views equal influence, and loss-based selection ties allocation to a moving model state. Voxel spacing provides a difficulty order before the model makes a prediction.

LatticeRank turns that order into supervision: endpoint-matched candidates share one network during training, their posteriors return to the reference lattice, and disagreement is measured against a detached reference prediction, with only the candidate of largest requested spacing supplying the auxiliary consistency loss. The training structure then disappears---inference is one direct-grid forward, with no candidate branch or ensemble. On Task04, the controlled ladder progresses from grid exposure to aligned consistency and finally to spacing-ranked allocation; the last step improves Extra Dice over mean consistency by 0.03436 across five seeds while meeting the prespecified reference-grid noninferiority criterion.

Our contributions are threefold: we formulate native-grid segmentation across voxel spacings as an ordered-view learning problem rather than exchangeable augmentations; we introduce a physically aligned, spacing-ranked consistency objective that concentrates supervision on the coarsest candidate without changing inference; and a controlled training ladder separates grid exposure, posterior consistency, and loss allocation, quantifying the accuracy--efficiency boundary against restoration.

\begin{figure}[t]
  \centering
  \includegraphics[width=\textwidth]{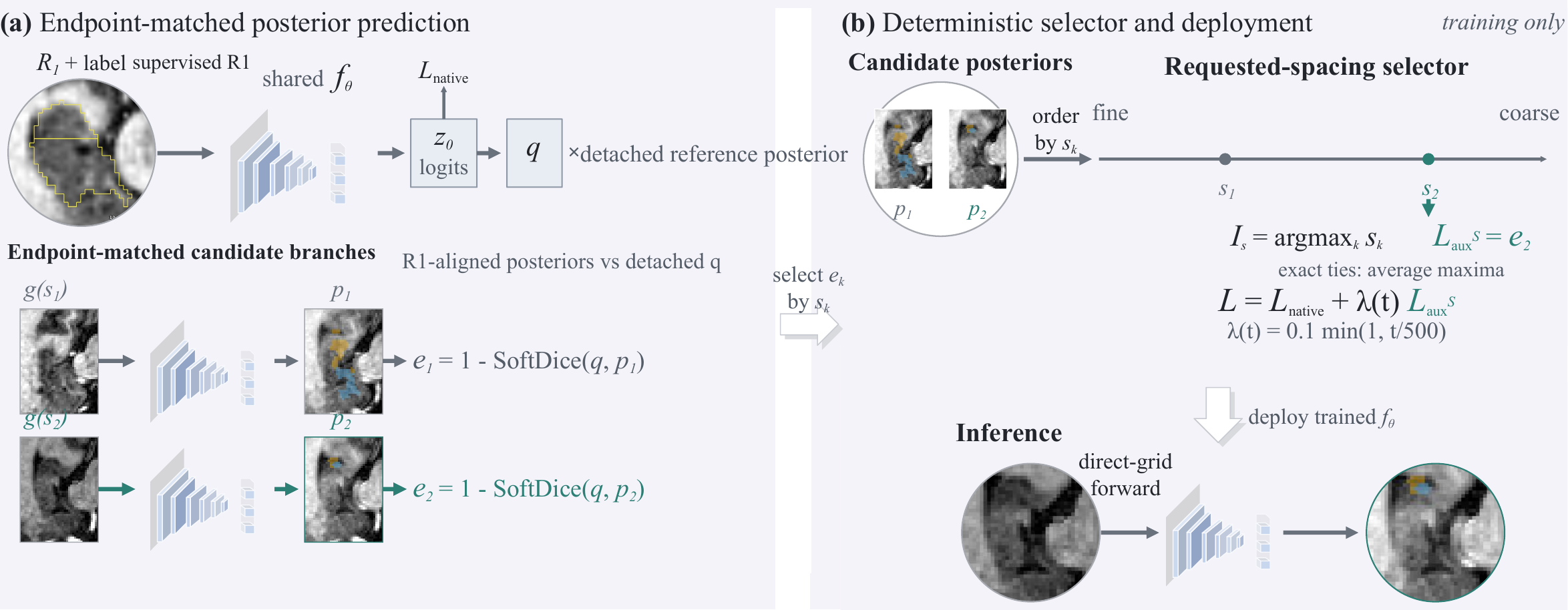}
  \caption{LatticeRank training and deployment. Endpoint-matched candidate grids share the segmentation network, and their posteriors are aligned to R1 before consistency is measured against a detached reference posterior. Requested spacing orders the candidates from fine to coarse; the coarsest candidate supplies the auxiliary loss. Candidate branches and alignment operations are removed at inference, leaving one direct-grid forward.}
  \label{fig:method}
\end{figure}

\section{Related Work}

\subsection{Resolution harmonization and spacing-adaptive segmentation}

Work on voxel spacing can be read by where it intervenes. At the input, nnU-Net estimates a target spacing from dataset statistics~\cite{isensee2021nnunet}, SynthSeg trains across synthetic contrasts and resolutions~\cite{billot2023synthseg}, and Consispace examines prediction stability under spacing changes~\cite{you2026consispace}. Representation- and architecture-level methods move the intervention inside the model: Splatting maps variable-resolution inputs into a shared space~\cite{brudfors2022splatting}, FastSurferVINN normalizes resolution internally~\cite{henschel2022fastsurfervinn}, HyperSpace generates spacing-conditioned weights~\cite{joutard2024hyperspace}, and CROWn addresses aliasing and phase mismatch in 3D downsampling and fusion~\cite{huang2026crown}. These approaches establish spacing as a first-class design variable; LatticeRank uses it to allocate consistency loss during training, with the interpolation operator and backbone held fixed.

\subsection{Consistency learning across aligned views}

Consistency learning offers a different route: preserve predictions while the input, teacher, task, or decoder changes. Mean Teacher introduced weight-averaged teacher targets~\cite{tarvainen2017meanteacher}; medical segmentation work extended the idea through transformation consistency, uncertainty-aware self-ensembling, and teacher--student perturbations~\cite{bortsova2019consistency,yu2019uamt,li2021tcsm}. Dual-task, cross-decoder, and cross-network objectives obtain complementary predictions in other ways~\cite{luo2021dtc,ouali2020cct,chen2021cps}, and recent work connects global and local representations across resolutions~\cite{liu2026crgloco}. The recurring question is how to construct a reliable target; here the targets are already defined after physical alignment, and the unresolved decision is how much optimization weight each ordered disagreement should receive.

\subsection{Ordered, hard, and worst-view learning}

Once several candidate losses coexist, the problem becomes one of selective optimization. Curriculum and self-paced learning schedule examples by difficulty~\cite{bengio2009curriculum,kumar2010selfpaced}; online hard-example mining and focal loss increase the influence of difficult samples or predictions~\cite{shrivastava2016ohem,lin2017focal}; MaxUp backpropagates through the highest-loss augmentation among sampled views~\cite{gong2021maxup}; and AdvChain and ASC learn challenging medical-image transformations or augmentation severities~\cite{chen2022advchain,alamdar2026asc}. LatticeRank uses a geometric ordering of views: requested spacing determines the selection before prediction, and physical alignment makes the posterior disagreements comparable. Unlike loss-based hard-view selection, this rule is independent of the current model state.

\subsection{Degradation-specific enhancement and medical reconstruction}

Degradation-specific networks accept the input as given: shadow removal via mask-guided routing~\cite{zeng2026dynroutenet} or multi-stage benchmarks~\cite{11462271}, underwater color fidelity~\cite{chen2026aquaclarity}, document contamination~\cite{chen2026docunfold}, and cinematic filter emulation from controlled datasets~\cite{11460866}; high-frequency synthesis under adverse acquisition~\cite{11523523} and compact compression of high-dimensional outputs~\cite{11609274}. Medical deployment raises the stakes: robotic surgical guidance enhances degraded camera feeds~\cite{li2026rga-net}, functional ultrasound is reconstructed auto-regressively~\cite{11297154}, generative models construct brain networks from diffusion imaging~\cite{11098517}, and image synthesis is giving way to digital-twin medical computing~\cite{chen2026image}. These systems pay off when the target acquisition is known in advance; LatticeRank instead varies one acquisition parameter continuously.

\section{Method}

\subsection{Ordered native grids}

Let $x^{(1)}$ and $y^{(1)}$ denote an image and label on the reference lattice R1 with nominal through-plane spacing 1~mm. For a requested spacing $s$, an endpoint-matched grid $g(s)$ preserves the physical upper and lower slice endpoints while changing the integer depth. Images are resampled directly to $g(s)$; only the through-plane spacing varies, while in-plane spacing remains 1~mm. The same network $f_\theta$ processes R1 and all candidate tensors, exposing the model to the discrete geometries of direct-grid inference without a spacing-specific backbone.

The candidate set $\{g(s_k)\}_{k=1}^{K}$ is ordered by requested spacing: larger $s_k$ yields fewer through-plane samples and, under endpoint matching, a coarser representation of the same physical extent. This order is a geometric severity prior, independent of labels, confidence, prediction error, gradients, or model state.

This order is deliberately defined on the requested spacing rather than an observed loss: endpoint quantization can make two requests realize the same integer depth, but the requested value remains the frozen variable that generated the view, supplying a deterministic allocation rule. The order encodes the expected information loss of coarser sampling, while posterior disagreement captures the case-specific response.

\subsection{Physically aligned posterior consistency}

The R1 branch produces logits $z_0=f_\theta(x^{(1)})$ and a detached foreground posterior
\begin{equation}
q=\operatorname{stopgrad}\!\left[\operatorname{softmax}(z_0)\right].
\end{equation}
For candidate $k$, the network prediction is mapped back to R1 with the known endpoint-matched transform:
\begin{equation}
p_k=\operatorname{Align}_{g(s_k)\rightarrow \Rone}\!\left[\operatorname{softmax}(f_\theta(x^{g(s_k)}))\right].
\end{equation}
Alignment is essential: disagreement must compare corresponding physical locations, not raw tensor indices. We define the candidate disagreement by foreground soft Dice,
\begin{equation}
e_k=1-\operatorname{SoftDice}(q,p_k).
\end{equation}
The uniform consistency comparator averages all candidate disagreements,
\begin{equation}
L_{\mathrm{aux}}^{C}=\frac{1}{K}\sum_{k=1}^{K}e_k.
\end{equation}

\subsection{Spacing-ranked allocation}

Uniform consistency treats candidate grids as exchangeable. LatticeRank instead selects the maximal requested spacing,
\begin{equation}
I_s=\arg\max_k s_k, \qquad
L_{\mathrm{aux}}^{S}=\frac{1}{|I_s|}\sum_{k\in I_s} e_k,
\end{equation}
where exact requested-spacing ties are averaged. In the main method, $K=2$ and $s_k$ is drawn from a frozen log-uniform schedule on $[1,3]$~mm. Selecting the maximum shifts the distribution of views contributing auxiliary gradient toward coarser grids. Unlike a worst-loss rule, selection cannot switch when the network temporarily mispredicts a mild view; the allocation remains attached to the known grid geometry.

The supervised R1 objective and total training loss are
\begin{align}
L_{\mathrm{native}}&=0.5L_{\mathrm{Dice}}(z_0,y^{(1)})+0.5L_{\mathrm{CE}}(z_0,y^{(1)}),\\
L&=L_{\mathrm{native}}+\lambda(t)L_{\mathrm{aux}},\qquad
\lambda(t)=0.1\min(1,t/500),
\end{align}
where $t$ is the update index. As illustrated in \cref{fig:method}, the reference and candidate branches exist only during training; inference receives one tensor on its deployment grid, executes $f_\theta$ once, and maps the posterior to R1 only when a common-grid output is required.

\section{Experiments}

\subsection{Dataset and implementation}

Experiments use MSD Task04 Hippocampus MRI with anterior and posterior hippocampal labels~\cite{antonelli2022msd}; the disjoint groups contain 80 training, 20 validation, 20 development, and 40 held-out cases. The main backbone is a compact Static Mamba implementation with 99,411 parameters, used as implementation context rather than the contribution~\cite{gu2024mamba}; a same-dataset development analysis replaces only the backbone with a plain 3D U-Net~\cite{cicek2016unet} (\cref{sec:sensitivity}).

All main runs use full volumes, batch size one, AdamW with learning rate $3\times10^{-4}$ and weight decay $10^{-5}$, bfloat16, no additional augmentation, and 2,500 updates; validation occurs every 250 updates, and the checkpoint with the best R1 foreground Dice is retained. The full ladder uses seeds 17/29/41 to separate the contributions of exposure, consistency, and allocation; the primary S-versus-C comparison adds seeds 53/67 for five training runs.

\subsection{Grid construction and optimization controls}

Candidate requests are drawn from frozen per-seed log-uniform schedules, and each candidate image is resampled directly from the raw volume to its endpoint-matched lattice rather than generated recursively from R1. Candidate probabilities return to R1 by the known affine transform with \texttt{align\_corners} disabled; the corresponding label is used only by DirectLabel, while C, D, and S compare aligned candidate posteriors with the detached R1 posterior. This common construction prevents interpolation chains or mismatched physical coordinates from becoming hidden method differences.

C, D, and S all construct and forward two candidate tensors per update; C and D backpropagate through both disagreements, whereas S backpropagates through only the spacing-selected one. The backbone contains LayerNorm and GroupNorm but no BatchNorm, dropout, running-statistic buffer, or other state-updating layer, so the unselected S forward is unnecessary for the objective or persistent state; it is retained only to match the two candidate forwards in C and D. Random single-candidate allocation is gradient-matched to S but not forward-matched, hence a selector control rather than a compute-equivalent baseline.

\subsection{Controlled training ladder}

The ladder separates the source of direct-grid improvement. \textbf{Single} trains only on R1 and is forwarded directly at every evaluation spacing; \textbf{Restore} resamples each coarse tensor back to R1 before model input; \textbf{DirectLabel} forwards candidate tensors directly and supervises their R1-mapped predictions with mapped labels; \textbf{C} replaces candidate labels with mean aligned-posterior consistency; \textbf{D} weights disagreements with detached error-softmax weights; and \textbf{S}, the proposed LatticeRank rule, assigns consistency to the candidate with maximal requested spacing. Random single-candidate allocation, soft spacing weighting, realized-spacing selection, candidate-count changes, and training-range changes are reported in \cref{sec:mechanism,sec:sensitivity}.

\subsection{Evaluation and statistics}

Evaluation uses requested spacings $[1,1.5,2,2.5,3,4,5]$~mm; because training samples candidates only from 1--3~mm, we designate 4 and 5~mm as the out-of-range coarse-grid stress regime. Their mean mapped Dice was fixed as \textbf{Extra Dice} before the five-seed S-versus-C comparison, and the complete spacing curves are reported to avoid selecting favorable endpoints. $\Dmap(s)$ compares a candidate prediction mapped to R1 with the R1 label; \textbf{Worst-grid Dice} is the minimum over 1.5--5~mm; Grid-AUC integrates prediction agreement $\Dpred$ over 1.5--5~mm and is not an accuracy metric.

Paired contrasts use a crossed bootstrap with 10,000 draws, independently resampling cases and training seeds before averaging their crossed product; the primary comparison is S minus C Extra Dice on the 40 held-out cases and five seeds. R1 Dice is monitored with a $-0.005$ noninferiority margin. The deployment benchmark measures warm-cache operator latency from the same precomputed coarse tensor to the same R1 hard prediction, excluding image I/O and raw-to-coarse generation.
\section{Results}

\subsection{Grid exposure and consistency recover direct-grid segmentation}

\begin{figure}[t]
  \centering
  \includegraphics[width=\textwidth]{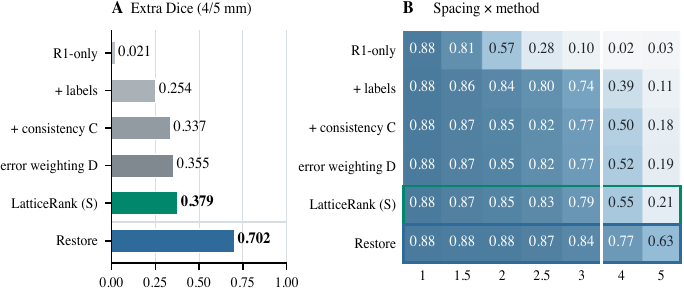}
  \caption{Controlled ladder on 40 held-out Task04 cases and seeds 17/29/41. Candidate exposure supplies the largest direct-grid recovery, aligned posterior consistency adds a second gain, and LatticeRank (S) achieves the strongest coarse-grid result among direct methods. Restoration remains the accuracy-oriented path.}
  \label{fig:ladder}
\end{figure}

The complete ladder reveals where performance is gained (\cref{fig:ladder} and \cref{tab:ladder}). A model trained only on R1 nearly collapses at the 4/5-mm stress grids, with Extra Dice 0.02102. Direct candidate exposure with mapped-label supervision raises this value to 0.25414, a gain of 0.23312 [0.19043, 0.27650]. Replacing mapped labels with aligned posterior consistency produces a second step: C reaches 0.33715, improving on DirectLabel by 0.08301 [0.06975, 0.09473], and D and S further increase Extra Dice to 0.35523 and 0.37855.

\begin{table}[t]
\caption{Held-out ladder. Extra is mean mapped Dice at 4/5~mm; Worst is the minimum mapped Dice over 1.5--5~mm. Grid-AUC measures agreement with the R1 prediction. For Extra, Worst, and Grid-AUC, best values are bold and second-best values are underlined.}
\label{tab:ladder}
\centering
\small
\setlength{\tabcolsep}{4.2pt}
\begin{tabular}{@{}lrrrrr@{}}
\toprule
Method & R1 & Extra & Worst & Grid-AUC & Train h \\
\midrule
Single & 0.88169 & 0.02102 & 0.00885 & 0.21639 & 0.63 \\
Restore & 0.88305 & \textbf{0.70226} & \textbf{0.63056} & \textbf{0.86056} & 2.80 \\
DirectLabel & 0.88348 & 0.25414 & 0.11299 & 0.60313 & 2.69 \\
C & 0.88365 & 0.33715 & 0.17547 & 0.65277 & 1.85 \\
D & 0.88417 & 0.35523 & 0.18959 & 0.66329 & 1.84 \\
S (LatticeRank) & 0.88344 & \underline{0.37855} & \underline{0.20926} & \underline{0.68032} & 1.60 \\
\bottomrule
\end{tabular}
\end{table}

\subsection{Spacing ranking improves over uniform consistency}

\begin{figure}[ht]
  \centering
  \includegraphics[width=\textwidth]{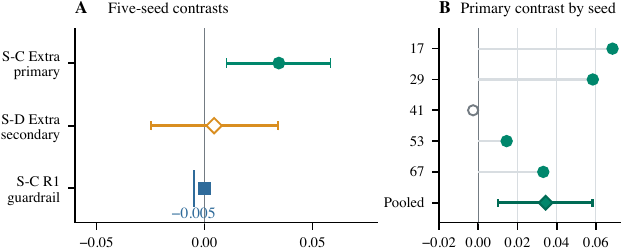}
  \caption{Five-seed S-versus-C evidence. LatticeRank improves Extra Dice over uniform consistency, while the R1 difference remains above the prespecified noninferiority margin. Seed-level effects show the variation across training runs, including the negative effect for seed 41.}
  \label{fig:primary}
\end{figure}

Across five seeds, LatticeRank improves Extra Dice over mean consistency by 0.03436, with a crossed 95\% confidence interval of [0.01013, 0.05824] (\cref{fig:primary}). Four of the five seed-level effects are positive; seed 41 is retained in the negative direction. The R1 contrast is 0.00012 [${-}$0.00247, 0.00255], its lower bound above the $-0.005$ guardrail, so the reference-grid result meets the prespecified noninferiority criterion. The S-minus-D Extra Dice contrast is 0.00443 [${-}$0.02495, 0.03387]; this interval includes zero, and the five-seed evidence establishes the gain over uniform consistency.

\subsection{What the ranking rule changes}
\label{sec:mechanism}

The secondary controls compare spacing selection with random allocation and alternative spacing weights: a random single-candidate rule reaches Extra Dice 0.32289, giving S minus Random $+0.05566$ [0.04155, 0.07290]: selecting only one disagreement is insufficient by itself. Soft spacing weighting reaches 0.36262, and S improves on this formulation by 0.01593 [0.01178, 0.01918]. Requested- and realized-spacing selection choose the same strict winner at every recorded update, differing only when endpoint quantization creates an exact realized-spacing tie; the realized-spacing rule reaches 0.37533, only 0.00322 below S. These comparisons support coarse-grid emphasis among the tested rules.

\begin{figure}[t]
  \centering
  \includegraphics[width=\textwidth]{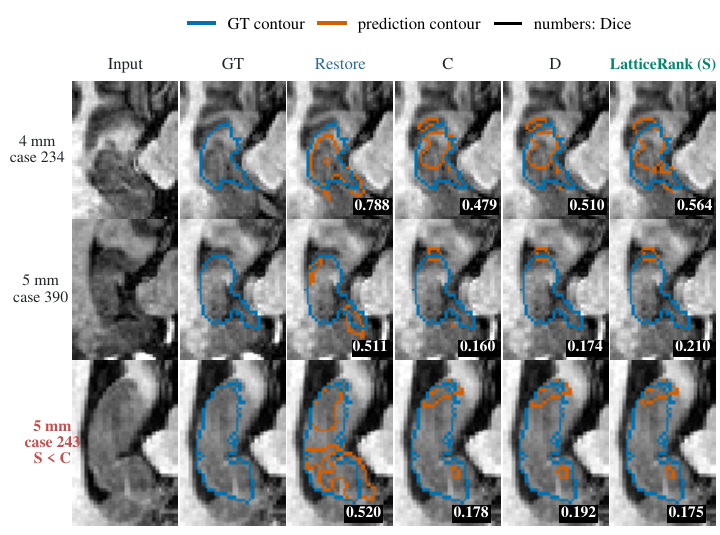}
  \caption{Unmodified seed-17 qualitative cases. Rows show a 4-mm case, a 5-mm median-bracketing case, and the lowest 5-mm $S-\max(C,D)$ case. Columns show input, ground truth, restoration, C, D, and LatticeRank (S). Foreground Dice (restore/C/D/S) is 0.788/0.479/0.510/0.564, 0.511/0.160/0.174/0.210, and 0.520/0.178/0.192/0.175 by row. Blue and orange denote ground-truth and prediction contours.}
  \label{fig:qualitative}
\end{figure}

The qualitative cases preserve the same message without claiming per-case dominance (\cref{fig:qualitative}): at 4~mm and in one 5-mm median-bracketing case, S produces the strongest direct-grid contour among C/D/S, while the final row shows the lowest 5-mm $S-\max(C,D)$ case, where S is below both comparators; coarse sampling leaves substantial information loss even with consistency training.

\subsection{Candidate-set sensitivity and backbone transfer}
\label{sec:sensitivity}

\begin{figure}[t]
  \centering
  \includegraphics[width=\textwidth]{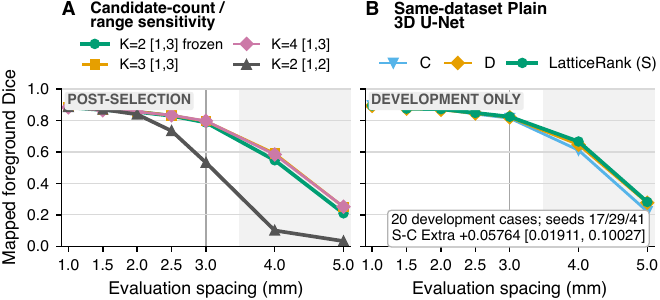}
  \caption{Candidate-set sensitivity and same-dataset backbone comparison. Left: post-selection analyses of candidate count and training range on the three-seed held-out panel. Right: C/D/S with a Plain 3D U-Net on the development set. These supporting analyses do not redefine the main configuration.}
  \label{fig:sensitivity}
\end{figure}

The frozen main configuration uses two candidates sampled from 1--3~mm; post-selection checks vary one aspect of this choice at a time (\cref{fig:sensitivity} and \cref{tab:sensitivity}). Increasing the candidate count to three raises Extra Dice from 0.37855 to 0.42039, while a fourth gives 0.41774, indicating no monotonic benefit beyond three candidates in this small panel; these variants also increase update time and peak memory and do not replace the frozen $K=2$ setting. Narrowing the training range to 1--2~mm reduces Extra Dice to 0.06705; the severe loss is concentrated outside the shortened range, indicating that the selector cannot compensate for missing coarse-grid exposure.

A separate Plain 3D U-Net development block asks whether the ordering signal depends on the Static Mamba implementation. With the backbone changed and the C/D/S protocol otherwise retained, S reaches Extra Dice 0.47420, compared with 0.41656 for C and 0.46165 for D; the $S-C$ contrast is $+0.05764$ [0.01911, 0.10027], with all three seed-level directions positive. This 20-case development check supports transfer of the ordering effect to a second backbone, although its R1 lower confidence bound slightly crosses the main $-0.005$ guardrail.

\begin{table}[t]
\caption{Candidate-set sensitivity and same-dataset backbone transfer. The upper block is post-selection held-out analysis; the lower block is development-only. Extra is mean mapped Dice at 4/5~mm.}
\label{tab:sensitivity}
\centering
\small
\setlength{\tabcolsep}{3.8pt}
\begin{tabular}{@{}llrrrr@{}}
\toprule
Setting & Rule & R1 & Extra & Worst & Grid-AUC \\
\midrule
$K=2$, 1--3~mm & S (frozen) & 0.88344 & 0.37855 & 0.20926 & 0.68032 \\
$K=3$, 1--3~mm & S & 0.88573 & 0.42039 & 0.25008 & 0.70160 \\
$K=4$, 1--3~mm & S & 0.88361 & 0.41774 & 0.25150 & 0.70080 \\
$K=2$, 1--2~mm & S & 0.88576 & 0.06705 & 0.02936 & 0.44836 \\
\midrule
Plain 3D U-Net & C & 0.89631 & 0.41656 & 0.22196 & 0.72072 \\
Plain 3D U-Net & D & 0.89268 & 0.46165 & 0.27688 & 0.73891 \\
Plain 3D U-Net & S & 0.89455 & 0.47420 & 0.28218 & 0.75036 \\
\bottomrule
\end{tabular}
\end{table}

\subsection{Accuracy--efficiency boundary}

\begin{figure}[t]
  \centering
  \includegraphics[width=\textwidth]{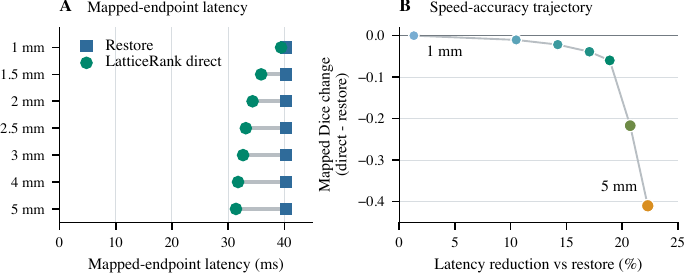}
  \caption{Common-endpoint operator benchmark. Direct forwarding reduces warm-cache latency relative to restoration, but the accuracy gap grows on coarse grids and no peak-memory advantage is observed.}
  \label{fig:deployment}
\end{figure}

LatticeRank targets pipelines that require inference on the available grid, improving direct-grid accuracy without inference branches. Restoration provides Extra Dice 0.70226, compared with 0.37855 for LatticeRank. From a precomputed coarse tensor to a common R1 hard prediction, its latency ratio is 0.850 [0.843, 0.857], roughly 15\% lower warm-cache latency (median 33.476~ms versus 40.270~ms for restoration), with peak allocated memory effectively unchanged. As spacing becomes coarser, direct forwarding becomes faster but loses more accuracy (\cref{fig:deployment}); restoration remains preferable when accuracy takes priority.

\section{Discussion}

Spacing supplies a stable prior over expected information loss. Mean consistency ignores that asymmetry and lets mild views dilute the coarse-grid signal; error- or uncertainty-based selectors can focus learning, but their notion of difficulty moves with the model. Requested spacing is available before optimization, and physical alignment keeps the selected disagreement anatomically meaningful, placing LatticeRank between resolution-aware segmentation and worst-view learning without altering the inference network. Compositional concept extraction~\cite{DBLP:conf/cvpr/ShiTHJCPWP26} offers one auditing route for what the network uses; the spacing ordering remains an inspectable geometric prior.

The evaluation uses controlled through-plane resampling of Task04 MRI, with 4/5~mm as out-of-range stress tests; it measures robustness under this construction, while clinical reliability at these spacings requires task-specific validation. Independent acquisition shifts, external datasets, and matched comparisons with spacing-adaptive systems remain directions for evaluation.

\section{Conclusion}

LatticeRank treats voxel-spacing candidates as ordered views rather than exchangeable augmentations. By aligning their posteriors in physical space and assigning consistency to the coarsest requested grid, it improves coarse-grid segmentation over uniform consistency while meeting the prespecified reference-grid noninferiority criterion and retaining single-forward inference. Restoration remains the stronger accuracy path, but LatticeRank moves direct-grid inference to a better point on the accuracy--latency curve; requested spacing thus provides a simple, model-independent training signal for direct-grid segmentation under the tested through-plane spacing changes.

\begin{credits}
\subsubsection{\ackname} This work was supported by the National Natural Science Foundation of China (Grant No. 62001184), the Guangdong Basic and Applied Basic Research Foundation (Grant No. 2025A1515110445), the Projects of Educational Commission of Guangdong Province (Grant Nos. 2025ZDZX1038, 2025KTSCX49, and 2023ZDZX1025), the Huizhou University Independent Innovation Capability Enhancement Plan Project (Grant No. HZU202535), and the College Students' Innovative Entrepreneurial Training Plan Program (Grant Nos. 202410577008 and 202610577034).
\end{credits}

\bibliographystyle{splncs04}
\bibliography{references}

\end{document}